\documentclass[letterpaper, 10 pt, conference]{ieeeconf}
\usepackage{times}
\usepackage[pdftex]{graphicx}
\usepackage{subfigure}
\usepackage{amsmath,amssymb,amsopn,amstext,amsfonts}
\usepackage{cancel}
\usepackage[space]{cite}
\usepackage{pdfsync}
\usepackage{balance}
\usepackage{color}
\usepackage{mathtools}
\usepackage{makecell}
\usepackage{algorithm}
\usepackage{algorithmic}
\usepackage{bm}

\usepackage{diagbox}
\usepackage{float}
\usepackage{epstopdf}
\usepackage{pifont}
\usepackage{multirow}
\usepackage{url}
\usepackage[linkcolor=black,citecolor=black,urlcolor=black,colorlinks=true]{hyperref}

\graphicspath{{../figure/}}
\DeclareGraphicsExtensions{.png,.jpg,.eps,.pdf}
\IEEEoverridecommandlockouts
\title{\LARGE \bf EVA-Planner: Environmental Adaptive Quadrotor Planning}
\author{Lun Quan$^{1,2,3}$, Zhiwei Zhang$^{1,2,3}$, Xingguang Zhong$^{1,2}$, Chao Xu$^{1,2}$, and Fei Gao$^{1,2}$
	\thanks{This work was supported by National Natural Science Foundation of China under Grant 62003299 and Grant 62088101}
	\thanks{$^{1}$State Key Laboratory of Industrial Control Technology, Institute of Cyber-Systems and Control, Zhejiang University, Hangzhou 310027, China.}
	\thanks{$^{2}$Huzhou Institute, Zhejiang University, Huzhou 313000, China.}
	\thanks{$^{3}$National Engineering Research Center for Industrial Automation (Ningbo Institute), Ningbo 315000, China.}
	\thanks{E-mail:{\tt\small \{lunquan,cxu,fgaoaa\}@zju.edu.cn}}
}

\begin{document}

\maketitle
\thispagestyle{empty}
\pagestyle{empty}

\begin{abstract}
The quadrotor is popularly used in challenging environments due to its superior agility and flexibility.
In these scenarios, trajectory planning plays a vital role in generating safe motions to avoid obstacles while ensuring flight smoothness. Although many works on quadrotor planning have been proposed, a research gap exists in incorporating self-adaptation into a planning framework to enable a drone to automatically fly slower in denser environments and increase its speed in a safer area. In this paper, we propose an environmental adaptive planner to adjust the flight aggressiveness effectively based on the obstacle distribution and quadrotor state. Firstly, we design an environmental adaptive safety aware method to assign the priority of the surrounding obstacles according to the environmental risk level and instantaneous motion tendency. Then, we apply it into a multi-layered model predictive contouring control (Multi-MPCC) framework to generate adaptive, safe, and dynamical feasible local trajectories. Extensive simulations and real-world experiments verify the efficiency and robustness of our planning framework. Benchmark comparison also shows superior performances of our method with another advanced environmental adaptive planning algorithm. 
Moreover, we release our planning framework as open-source ros-packages\footnote{\label{web}https://github.com/ZJU-FAST-Lab/EVA-planner}.
\end{abstract}

\section{Introduction}
\label{sec:introduction}
Nowadays, quadrotors are increasingly used in dangerous scenarios such as mine exploration, quick response rescue, and target search~\cite{letquan2020}.
These applications may be difficult and dangerous for human beings and have high demands on the autonomy and robustness of drone. 

\begin{figure}[t]
	\begin{center}
		\includegraphics[width=1.0\columnwidth]{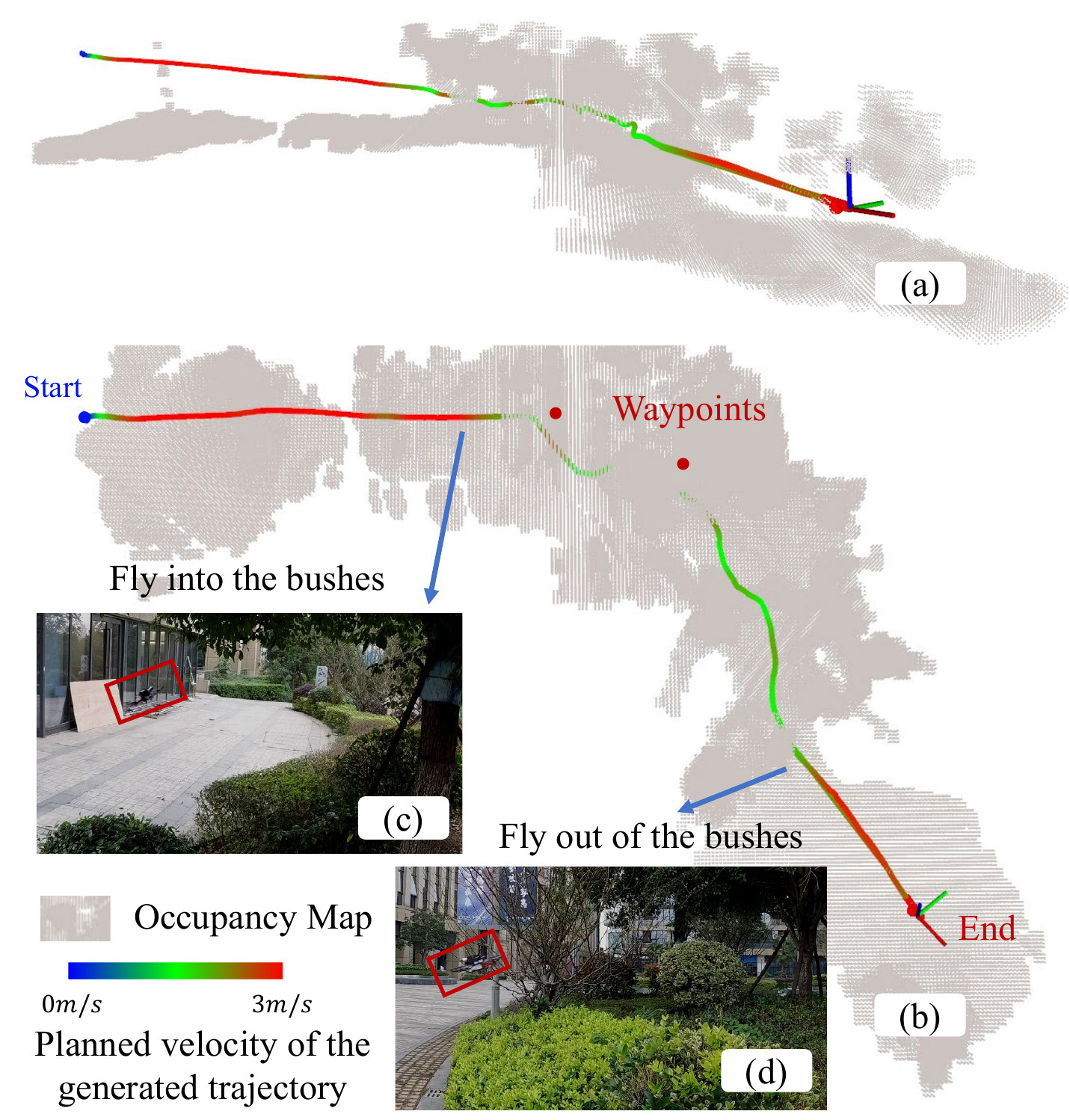}
	\end{center}
	\caption{(a) and (b) are the side and vertical views of the outdoor experiment. (c) and (d) are the snapshots of critical moments. The curve represents the generated trajectory, and the gradient color is the planned velocity. This experiment is described in detail in Sec.~\ref{sec:real_world}. Video is available at \url{https://www.youtube.com/watch?v=HcwBNcah0eo}.
		\label{fig:real}}
	\vspace{-0.5cm}
\end{figure}

Although there are extensive works on quadrotor motion planning, few of them consider the self-adaptation of flight aggressiveness.
Imagine this situation, for a quadrotor flies in a forest, as it moves more aggressively, there may occur more deviations in the state estimation, perception, and control. 
These errors result in a higher probability of a crash, especially while the quadrotor navigates between multiple obstacles in a narrow space. 
In contrast, if the quadrotor operates in a wide-open space, a large safety margin naturally promises that the drone can fly faster.
Moreover, in this scenario, the quadrotor should fly fast instead of limit its speed conservatively.
Based on the above analysis, an ideal planner needs to leave sufficient planning margin by automatically adjusting its aggressiveness according to the degree of risk, which is directly decided by the obstacle distribution and quadrotor dynamics. 

To this end, we propose an environmental adaptive quadrotor planning method, named EVA (\textbf{E}n\textbf{V}ironmental \textbf{A}daptive)-Planner, which effectively adjusts the flight aggressiveness based on multi-layered adaptive planning.
Our method takes the environmental risk level and instantaneous motion tendency into account. 
In this way, quadrotor can adaptively assign the priority of obstacles around it and plan a more safe and efficient flight trajectory. 
This method is implemented with a unified planning framework extending our recent results on quadrotor model predictive contouring control~\cite{ji2020cmpcc}. 
We summarize our contributions as: 

\begin{enumerate}
	\item An environmental adaptive safety aware method is proposed with a reasonable judgment of the danger degree of surrounding obstacles according to the environmental information and the system's motion tendency. 
	\item A unified planning framework for generating aggressiveness-adaptive, obstacle-free, smooth, and dynamics feasible trajectory using multi-layered model predictive contouring control.
	\item Extensive simulations and real-world experiments verify the robustness of our method. Benchmark shows the efficiency of our method in generating fast and safe trajectories.
\end{enumerate}

The rest of the paper is organized as follows, related works are discussed in Sec.\ref{sec:related_works}. Our method for environmental adaptive safety aware is stated in Sec.\ref{sec:adaptive}, and the multi-layered planning framework is described in detailed in Sec.\ref{sec:framework}. Simulation and real-world experiments results are discussed in Sec.\ref{sec:result}. This paper is concluded in Sec.\ref{sec:conclusion}.

\section{Related works}
\label{sec:related_works}

\subsection{Autonomous navigation algorithms}
\label{sec:auto_navigation}
The problem of online autonomous navigation in the 3D complex environment has been investigated extensively. According to the planner structure, the planning algorithms can be divided into direct and hierarchical methods.

Direct methods plan trajectories on the abstract environment, which reduces the computational load.
By establishing a safe flight corridor (SFC) directly on the depth map~\cite{LopezICRA2017} or point clouds~\cite{GaoJFR2019}, trajectory planning can be carried out without building the grid map. \textup{Nanomap}~\cite{nanomap2018} generates rough k-d trees for collision check. However, direct methods tend to fail in a complex environment due to the memoryless of the map. Hierarchical methods such as~\cite{ratliff2009chomp,oleynikova2016continuous,fei2017iros,zhou2019robust} employ a planning framework with front-end path searching and back-end trajectory optimizing, which transforms the trajectory generation to a nonlinear optimization problem and formulates it as trading off smoothness, obstacle-free, and dynamical feasibility simultaneously. However, the executed trajectory may hit the obstacles due to the tracking errors caused by the high-speed motion, even if the desired obstacle-free trajectory is planned. 

\subsection{Adaptive fast and safe planning}
\label{sec:fast_safe}
\textup{FASTER}~\cite{tordesillas2019faster} plans a safe trajectory and a fast trajectory simultaneously at each time and realizes safe-fast flight through practical replanning design. However, \textup{FASTER} uses mixed integer programming and can only maintain real time planning performance when optimizing in several polyhedras.
Some works mention the importance of jointly considering the geometry of configuration space and system dynamics, but few have shown convincing performance for a quadrotor in 3D dense environments. 
A planning framework is proposed in~\cite{fk2018icra} with offline precomputed safety margins relative to different system dynamics and an online generated tree of trajectories that adaptively search a proper planning model. 
However, if the quadrotor needs to fly at high speed, substantial pre-computed models prevent it from being searched online.
A new state-dependent directional metric is designed in~\cite{li2020icra} to adaptively adjust the influence of the environment on the system according to the velocity direction of vehicle.
This metric checks collision based on how dangerous an obstacle is.
In this way, obstacles parallel with velocity are considered more likely to cause collision than obstacles perpendicular to the velocity.
However, this method treats the direction along and reverse with the velocity as same, thus behaves rather conservative.
For instance, even when the vehicle is escaping from an obstacle-rich area, it also lowers its speed to satisfy the collision checker.

The above planning methods only consider the influence of the closest distance between the trajectory and the obstacles when dealing with safety, which is counterintuitive because the safety is also related to the system dynamics. For example, the degree of danger to the system is different when the quadrotor is flying towards and away from the obstacles, even if the distance is the same between the quadrotor and obstacles. In order to solve this problem, an environmental adaptive safety aware method is proposed in Sec.\ref{sec:adaptive}.

\section{Environmental adaptive safety aware}
\label{sec:adaptive}

\begin{figure}[t]
	\begin{center}
		\includegraphics[width=0.8\columnwidth]{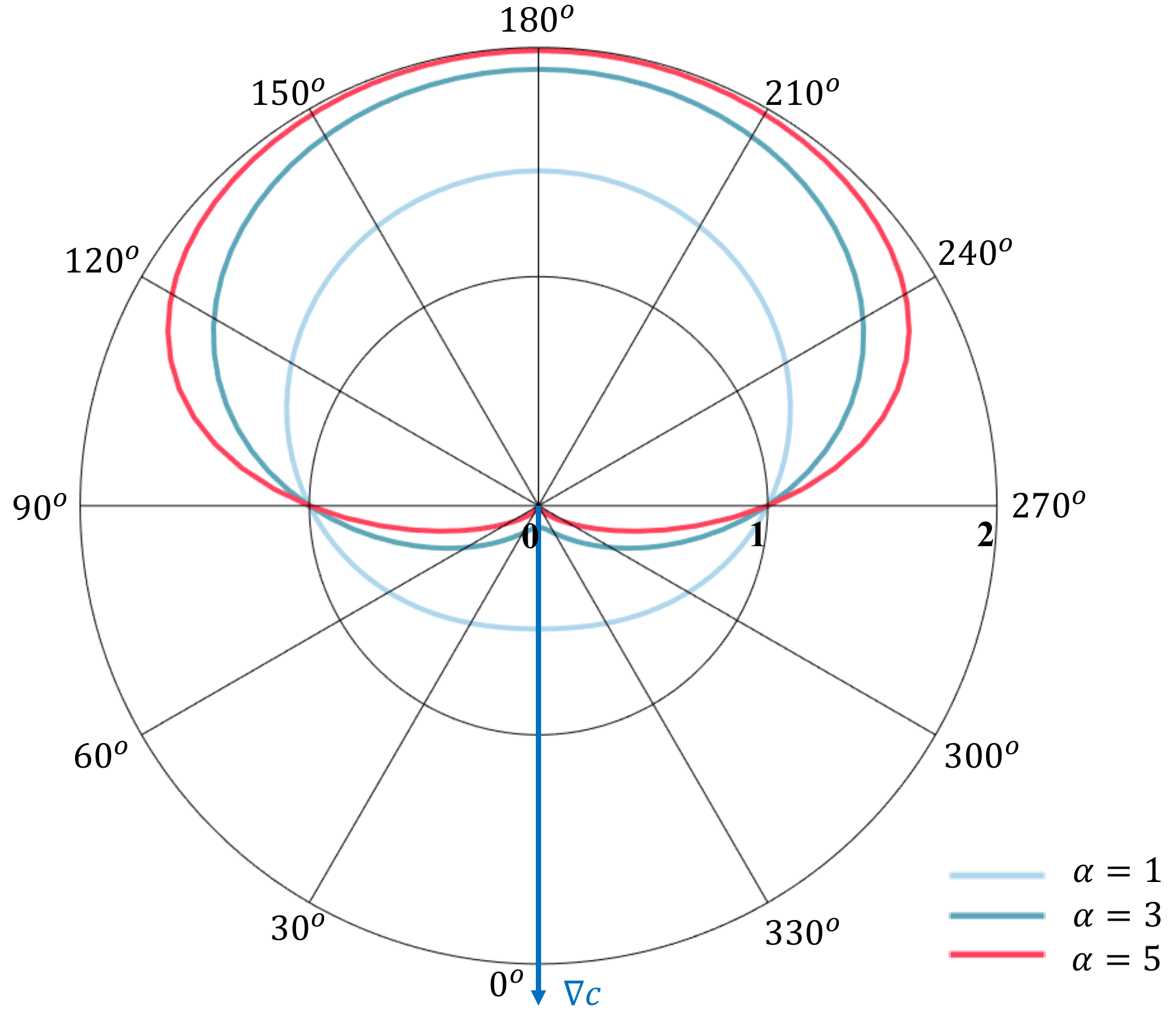}
	\end{center}
	\caption{The polar graph of $\eta$ with different $v$ when $\nabla c$ is constant to $(0,-1)$ in the 2D situation. The angle of the polar graph is between $v$ and $\nabla c$, and the length of the curve along the polar axis is the magnitude of $\eta$ corresponds to velocity direction.
		\label{fig:adaptive_risk}}
	\vspace{-0.5cm}
\end{figure}

As mentioned above, it is essential to adaptively adjust the safety aware of quadrotor according to the environmental risk level and instantaneous motion tendency. 
EVA-planner uses the environmental adaptive safety aware (EASA) to calculate the risk weight $\eta$ which regulates the flight behavior of quadrotor, which comprehensively considers the gradient $\nabla c$ of the Euclidean signed distance field (ESDF) with respect to the velocity $v$. 

Inspired by~\cite{li2020icra}, we realize that it is essential to assign the priority of obstacles depending on the system velocity. So we propose a method to adjust the priority of obstacles more intuitively than the above method, which uses the sigmoid function to map the cosine of the angle between $v$ and $\nabla c$ to the risk weight $\eta\in[0,2]$. The function is defined as
\begin{equation}
\label{eq：risk_weight}
	\eta(\beta)=\frac{2}{1+e^{\alpha \beta}},
\end{equation}
\begin{equation}
\label{eq：beta}
\beta=\frac{<v,\nabla c>}{\|v\| \|\nabla c\|},
\end{equation}
where $\alpha \in\mathbb{R}^+$ is the change rate coefficient and $<,>$ is the dot product operation.

As shown in Fig.\ref{fig:adaptive_risk}, $\eta$ is maximum when $v$ and $\nabla c$ are opposite, and $\eta$ is minimum when $v$ and $\nabla c$ are in the same direction. We simply regard the situation $\beta\in(0,1]$ as safe and the situation $\beta\in[-1,0]$ as dangerous. The impact of $\alpha$ in (\ref{eq：risk_weight}) is shown in the Fig.\ref{fig:adaptive_risk}. With the $\alpha$ increasing, the difference of the risk weight $\eta$ between dangerous and safe situations also increases. 

\begin{figure}[t]
	\begin{center}
		\includegraphics[width=1.0\columnwidth]{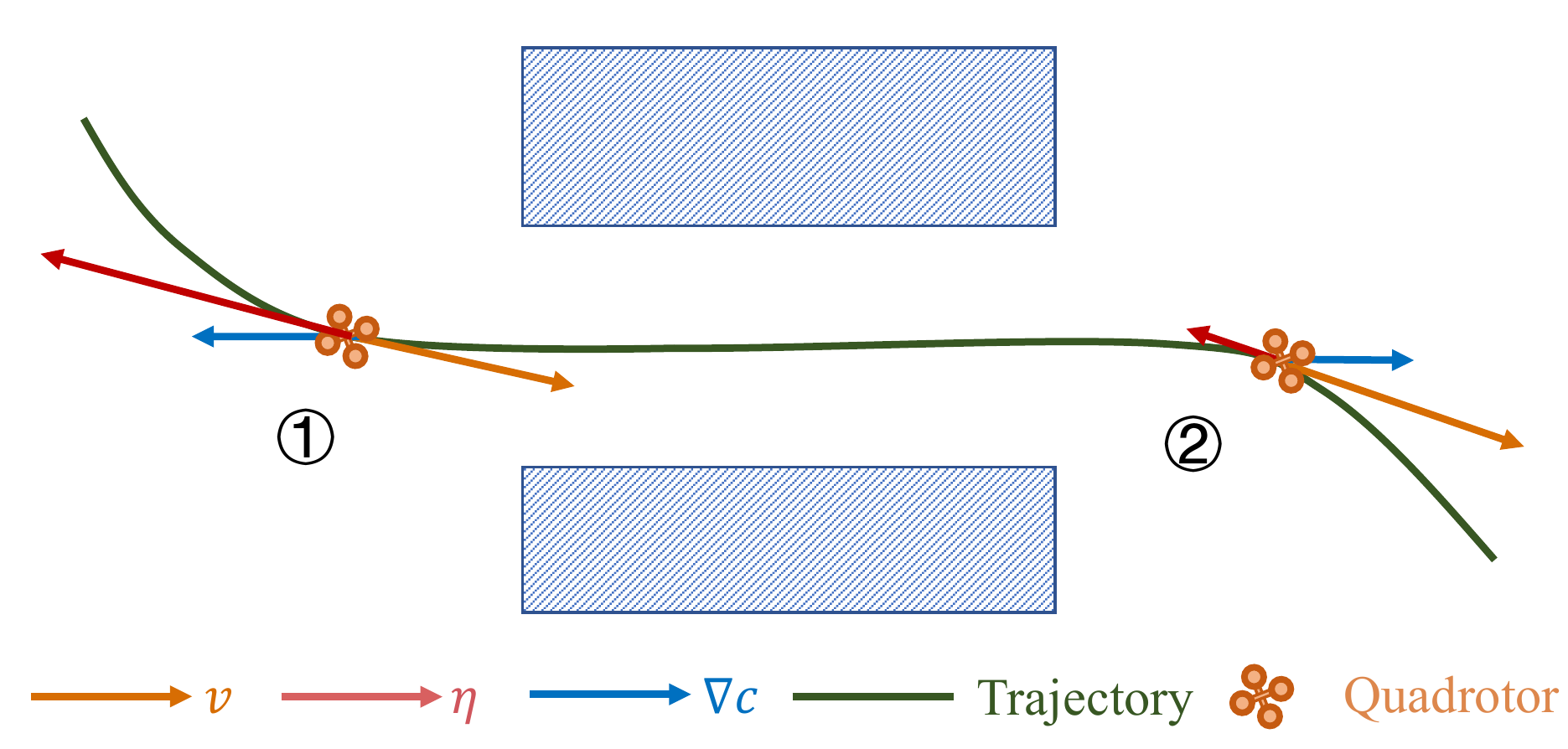}
	\end{center}
	\caption{Illustration of $\eta$ changing in the planning process. The risk weight $\eta$ is a scalar, but we put it in the opposite direction of $v$ to show its slowing effect on the quadrotor. 
		\label{fig:example_risk}}
	\vspace{-0.8cm}
\end{figure}

EASA reasonably corresponds each pair of $\{v,\nabla c\}$ to a risk weight $\eta$. This method provides more intelligent environmental information for the planning process, reducing conservative planning, and providing clear signals in dangerous situations. As shown in Fig.\ref{fig:example_risk}, $\eta$ is large when a quadrotor flies into the corridor, which means it is in a dangerous situation and needs to fly cautiously at a slower velocity. Moreover, $\eta$ is small when a quadrotor flies away from the corridor, which means it is in a safe state and can fly more aggressively. The application of EASA to the trajectory generation is described in detailed in Sec.\ref{sec:highmpcc}.

\section{Multi-layered Model predictive contouring control framework}
\label{sec:framework}
\subsection{Multi-layered planning framework}
\label{sec:planning_framework}
Thanks to the differential flatness of the quadrotor dynamics stated in~\cite{MelKum1105}, we deal with the quadrotor as a particle model. In the planning process, the trajectories of the three axes $x,y,z$ are decoupled, and the yaw angle $\psi$ planning is carried out independently just like~\cite{ral2020zhou}. 

This paper designs a planning framework with three coarse-to-fine layers to generate receding-horizon local adaptive trajectories in real-time. The framework is unified among these layers and can be applied to quadrotors with different computing forces. This property enables planners to deal with approximations of system with arbitrary high order, depending on the hardware.

In the first layer, we use the path finding algorithm, such as sampling-based method RRT*~\cite{karaman2011} or searching-based method A*~\cite{intro2009mit} to generate a global guiding path which connects the start and end points in the free space. 
In the middle layer, we use the low-order motion model to optimize the geometry of guiding path, named Low-level MPC. By doing so, the computing resources required to generate a reference trajectory can be reduced. In the last layer, we generate the local trajectory by optimizing EASA, tracking error, flight aggressiveness, and dynamical feasibility with high order system, named High-level MPCC.

\subsection{System representation}
\label{sec:representation}
The same system representation is used in Sec.\ref{sec:lowmpc} and Sec.\ref{sec:highmpcc} with different system dimension and order. The different parameter selections can be found in Tab.\ref{tab:system}. The time interval between the adjacent state points is fixed as $\delta t$. We assume that the system input $u_i$ is constant in $\delta t$. Therefore the kinematic relationship between the two adjacent points can be expressed as

\begin{equation}
\label{eq:one_step}
\mathbf{s}_{i+1} = A_d \mathbf{s}_i + B_d u_i,
\end{equation}
where $A_d$ and $B_d$ are the state-transfer matrixes governed by the $d^{th}$-order integral model. Based on the above assumptions, state sequence $\{\mathbf{s}_i\}$ can be represented as a mapping of the input sequence $\{u_i\}$ by using the state transition function

\begin{equation}
\label{eq:system_transition}
\mathbf{S}_\mu = A \mathbf{U}_\mu + B \mathbf{s}_0,
\end{equation}

\begin{equation}
\label{eq:A}
A = \begin{pmatrix} 0 & \cdots\ &\cdots\ & \cdots\ & 0 \\
					B_d  & 0 & \cdots\ & \cdots\ & 0    \\
					A_d B_d & B_d & 0 & \cdots\ &0 \\
					\vdots & \vdots & \vdots & \ddots  & \vdots \\
					A_d^{H-1} B_d & A_d^{H-2} B_d & A_d^{H-3} B_d & \cdots\ & B_d,
	 \end{pmatrix},
\end{equation}

\begin{equation}
\label{eq:B}
B = (I,A_d,{A_d}^2 \cdots\ {A_d}^H)^T,
\end{equation}

where state vector $\mathbf{S}_\mu=[\mathbf{s}_0,\mathbf{s}_1\cdots \mathbf{s}_H]^T$ and input vector $\mathbf{U}_\mu=[u_1,u_2\cdots u_H]^T$.

From (\ref{eq:system_transition}), all system states can be represented by ${\mathbf{U}_\mu}$ when the initial state $\mathbf{s}_0$ is known. Therefore the decision variables of the trajectory optimization problem are reduced to ${\mathbf{U}_\mu}$. Meanwhile, we construct the trajectory generation as an unconstrained optimization problem by transforming the constraints into penalty terms. The reference trajectory generation and the local trajectory optimization are explained in detail in Sec.\ref{sec:lowmpc} and Sec.\ref{sec:highmpcc} respectively.

\begin{table}[t]
	\caption{system representation of the multi-layered mpcc}
	\begin{tabular}{m{1.4cm} m{2.8cm} m{2.8cm}}
		\hline
		\hline
		& \makecell[c]{Low-level MPC}& \makecell[c]{High-level MPCC}\\
		\hline
		\makecell*[c]{System dimension} & \makecell*[c]{$\mu\in\{ x,y,z\}$} & \makecell*[c]{$\mu\in\{ x,y,z,\theta\}$} \\
		\makecell*[c]{System order} & \makecell*[c]{$d=1$} & \makecell[c]{$d=3$} \\
		\makecell*[c]{prediction horizons} & \makecell*[c]{$H=M$} & \makecell*[c]{$H=N$} \\
		\makecell*[c]{State} & \makecell*[c]{$\mathbf{s}_i=p_i$ \\ $i=0,1...M$} & \makecell*[c]{$\mathbf{s}_i=[p_i,v_i,a_i]^T$ \\ $i=0,1...N$}\\
		\makecell*[c]{input} & \makecell*[c]{$u_i=v_i$ \\ $i=1,2...M$} & \makecell*[c]{$u_i=j_i$ \\ $i=1,2...N$} \\
		\makecell*[c]{transform matrix} & \makecell*[c]{$A_1=1$\\$B_1=\delta t$} & \makecell*[c]{$A_3 = \begin{pmatrix}
			1 & \delta t & \frac{\delta t^2}{2} \\
			0 & 1 & \delta t \\
			0 & 0 & 1
			\end{pmatrix}$ \\ $B_3 = [\frac{\delta t^3}{6},\frac{\delta t^2}{2},\delta t]^T$} \\
		\hline
		\hline
	\end{tabular}
\label{tab:system}
\end{table}

\subsection{Low-level MPC}
\label{sec:lowmpc}
The primary purpose of this layer is to optimize a geometrically continuous and obstacle-free reference trajectory quickly.
Inspired by~\cite{zhou2020raptor}, the guiding path attracts the trajectory to escape the local minimum, and ESDF pushes the trajectory to find the local optimum. In this paper, we choose the velocity as the input to the reference trajectory, as shown in Tab.\ref{tab:system}. The objective function is   

\begin{equation}
\label{eq:lowf}
	\min \kappa_1 J_s +\kappa_2 J_c + \kappa_3 J_u,
\end{equation}
where $J_s$ is the similarity penalty of the distance between the trajectory and the guiding path, $J_c$ is the collision cost, $J_u$ is the smoothness term. And $\kappa_1,\kappa_2,\kappa_3$ are the weights of these penalty items respectively

\begin{equation}
\label{eq:Js}
J_s=\sum_{i=1}^{M}\|p_i-g_i\|^2,
\end{equation}
\begin{equation}
\label{eq:Jc}
J_c=\sum_{i=1}^{M}F_c(c(p_i)),
\end{equation}
\begin{equation}
\label{eq:Ju}
J_u=\sum_{i=1}^{M-1}(u_{i+1}-u_i)^2,
\end{equation}
where $g_i$ is the $i^{th}$ uniformly distributed point of the guiding path and $c(p)$ is the distance of point $p$ in the ESDF. Because the input $u_i$ is constant in $\delta t$, the sum of squares of the difference between adjacent inputs is used to indicate the smoothness of trajectory in (\ref{eq:Ju}). $F_c(\cdot)$ is the penalty function of collision cost as

\begin{equation}
\label{eq:Fc}
F_c(c(p))=\begin{cases}
(c(p)-c_{thr})^2, & \mbox{if   } c(p)<c_{thr} \\
0,  & \mbox{if   } c(p)\ge c_{thr} 
\end{cases},
\end{equation}
where $c_{thr}$ is the safe distance threshold. The collision cost starts to rise rapidly when the distance from the obstacles is less than this threshold.

The optimized reference trajectory $\rho_\mu(\theta)$ is parameterized by $\theta \in [0,M\delta t]$, which is equal to the time parameter.

\subsection{High-level MPCC}
\label{sec:highmpcc}
After generating a reference trajectory, we take model predictive contouring control (MPCC) in the final layer. In this layer, we choose the jerk as the input and add the state $\{p_{\theta,i},v_{\theta,i},a_{\theta,i}\}$ of reference points which travels on the reference trajectory as another system dimension, as shown in Tab.\ref{tab:system}.

\begin{figure}[t]
	\begin{center}
		\includegraphics[width=1.0\columnwidth]{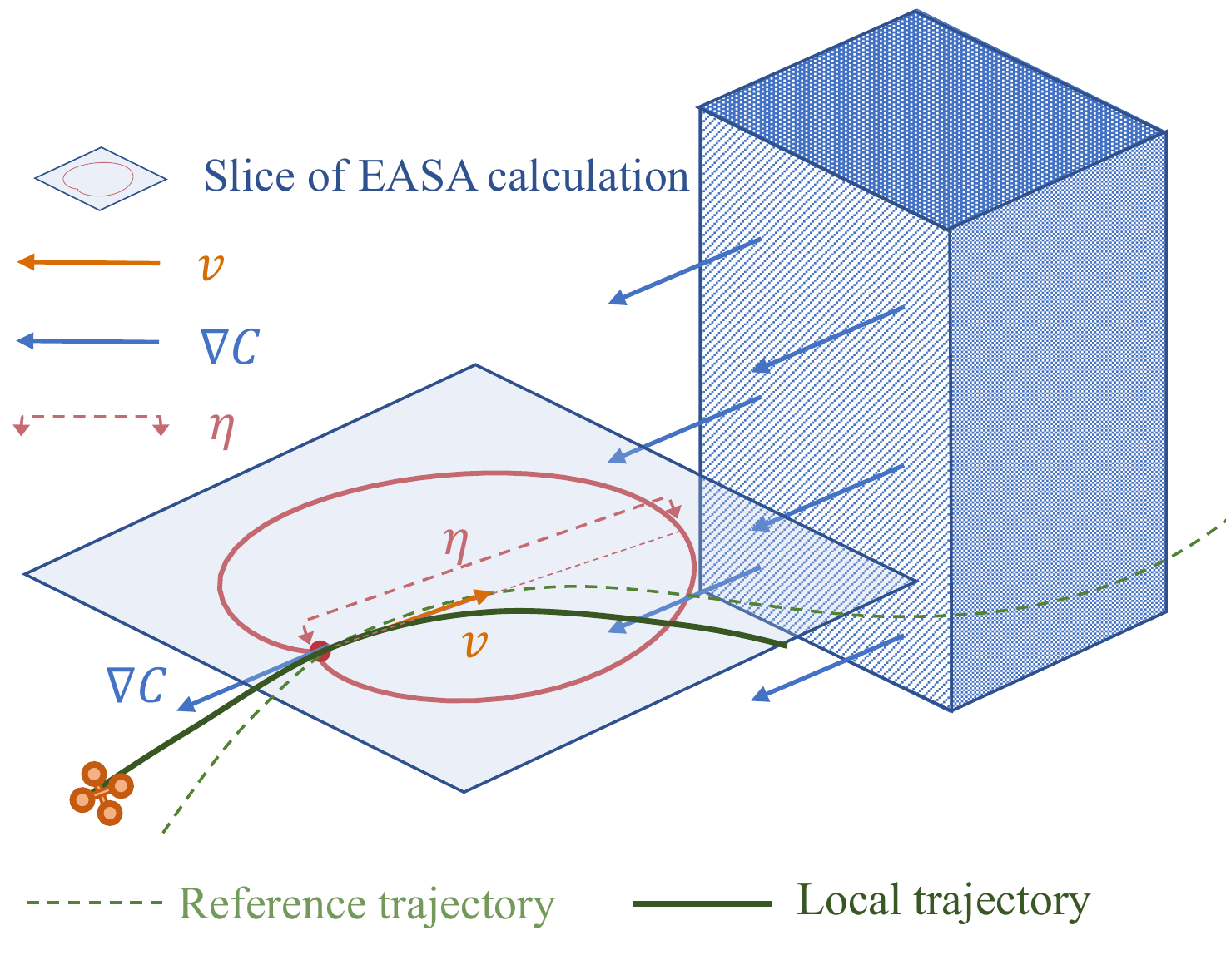}
	\end{center}
	\caption{Illustration of EASA calculation during trajectory optimization.
		\label{fig:adaptive}}
	\vspace{-0.3cm}
\end{figure}

Inspired by \cite{Liniger2015,ji2020cmpcc}, we optimize the tracking accuracy and traveling progress simultaneously. Meanwhile we consider the environmental adaptive flight by calculating EASA which is detailed in Sec.\ref{sec:adaptive}. Furthermore, there is also a penalty for violating the dynamical feasibility. The objective function is

\begin{equation}
\label{eq:highf}
\min \lambda_1 f_s +\lambda_2 f_p + \lambda_3 f_e + \lambda_4 f_c + \lambda_5 f_d,
\end{equation}
where $f_s$ is the tracking error between the local trajectory and the reference trajectroy, $f_p$ indicates the progress of reference points and $f_e$ represents EASA penalty itemm, $f_c$ is the collision cost same as (\ref{eq:Jc}) which is for obstacle avoidance and $f_d$ is the penalty for violating kinodynamics feasibility. $\lambda_i,i=1\cdots5$ are the weights of these items

\begin{equation}
\label{eq:fs}
f_s=\sum_{i=1}^{N}\|p_i-p_{\theta,i}\|^2,
\end{equation}

\begin{equation}
\label{eq:fp}
f_p=-\delta t\sum_{i=1}^{N} v_{\theta,i},
\end{equation} 

\begin{equation}
\label{eq:fe}
f_e =\sum_{i=1}^{N} \eta(\beta)F_c(c(p_i))(\|v_i\|-v_{thr})^2,
\end{equation}
above three cost function items are the main determinants of the trajectory behavior. The optimization problem trades off $f_s$ and $f_p$ to make the trajectory as fast as possible while ensuring the tracking error small enough. However, when the quadrotor flies towards obstacles, as shown in Fig.\ref{fig:adaptive}, $\eta$ increases to make $f_e$ become the primary determinant of the trajectory behavior, which causes the trajectory to slow down until the quadrotor is out of the dangerous area. The derivatives of (\ref{eq:fs}) and (\ref{eq:fp}) can be found in~\cite{Liniger2015}. The derivative of (\ref{eq:fe}) is

\begin{align}
\label{eq:fe_d}
\frac{\partial f_e}{\partial U_\mu} &= \sum_{i=1}^{N} \frac{2\alpha e^{\alpha \beta}}{(1+e^{\alpha \beta})^2} \frac{\partial \beta}{\partial U_\mu} F_c(c(p_i))(\|v_i\|-v_{thr})^2 \nonumber \\
&+ \eta(\beta) \nabla_\mu F_c(c(p_i)) \frac{\partial p_i}{\partial U_\mu} (\|v_i\|-v_{thr})^2 \nonumber 	\\
&+ \eta(\beta) 2F_c(c(p_i)) \frac{v_\mu}{\|v_i\|}(\|v_i\|-v_{thr}) \frac{\partial v_i}{\partial U_\mu}, \nonumber  \\
\frac{\partial \beta}{\partial U_\mu} &= \frac{\nabla_\mu c(p_i) \|v\|^2-v_\mu <v,\nabla c> }{\|v\|^3 \|\nabla c(p_i)\|} \frac{\partial v_i}{\partial U_\mu} \\
&+ \frac{v_\mu \|\nabla c(p_i)\|^2- \nabla_\mu<v,\nabla c> }{\|v\| \|\nabla c(p_i)\|^3} {\nabla_\mu}^2 c(p_i) \frac{\partial p_i}{\partial U_\mu}, \nonumber 
\end{align}
where ${\nabla_\mu}^2 c(p_i)$ is the second derivative in the third-order interpolation ESDF and $v_{thr}$ is generally set as $0.1m/s$ to prevent the optimized trajectory velocity from being zero. $\frac{\partial p_i}{\partial U_\mu}$ and $\frac{\partial v_i}{\partial U_\mu}$ are the corresponding row vectors of $A$ in (\ref{eq:A}). The obstacle avoidance and kinodynamics constraints are

\begin{equation}
\label{eq:fc}
f_c=\sum_{i=1}^{N}F_c(c(p_i)),
\end{equation}

\begin{equation}
\label{eq:fd}
f_d=\sum_{i=1}^{N} (F_d(v_i)+F_d(a_i)+F_d(j_i)+F_d(v_{\theta,i})),
\end{equation}
where $F_c(\cdot)$ is the penalty function of collision cost same as (\ref{eq:Fc}). $F_d(\cdot)$ is the penalty function of kinodynamics feasibility

\begin{equation}
\label{eq:Fd}
F_d(D)=\sum_{\mu} f_d(d_\mu),
\end{equation}

\begin{equation}
\label{eq:f_d}
f_d(d_\mu)=\begin{cases}
-(d_\mu-d_{min})^3, & d_\mu < d_{min}, \\
0,  & d_{min} \le d_\mu \le d_{max}, \\
(d_\mu-d_{max})^3, & d_\mu > d_{max},
\end{cases}
\end{equation}
where $d_{min}$ and $d_{max}$ are the lower and upper bound of each variable.

\section{Results}
\label{sec:result}
\subsection{Implementation Details}
\label{sec:implementation}
Extensive simulation and real-world flight experiments are carried out to test the effectiveness and robustness of proposed method. 
We use the flight platform proposed in~\cite{gao2019trr} with Intel Realsense D435\footnote{https://www.intelrealsense.com/depth-camera-d435/} for perception and mapping. Modules including state estimation, environment perception, trajectory planning, and flight control are all running onboard computer Manifold2\footnote{https://www.dji.com/cn/manifold-2} with Intel i7-8550U in real-time. All simulations run on a laptop with Intel i7-6500U.

In both simulation and real-world experiments, we generate the reference trajectory by solving (\ref{eq:lowf}) at a fixed interval of 2 seconds and check whether the reference trajectory has a collision with the new explored environment at a frequency of 100Hz. If the collision is detected, the reference trajectory replanning is carried out immediately. For local trajectory generation, we solve the unconstrained optimization problem (\ref{eq:highf}) with prediction horizons $N=40$ and time step $\delta t=0.05s$, so $T=2s$. The trajectory optimization problem is solved by a nonlinear optimization open-source library NLopt\footnote{https://nlopt.readthedocs.io/}. The optimization times of Low-level MPC and High-level MPCC are less than 10ms each time, respectively. 

\subsection{Benchmark}
\label{sec:benchmark}
\begin{figure}[h]
	\begin{center}
		\includegraphics[width=1.0\columnwidth]{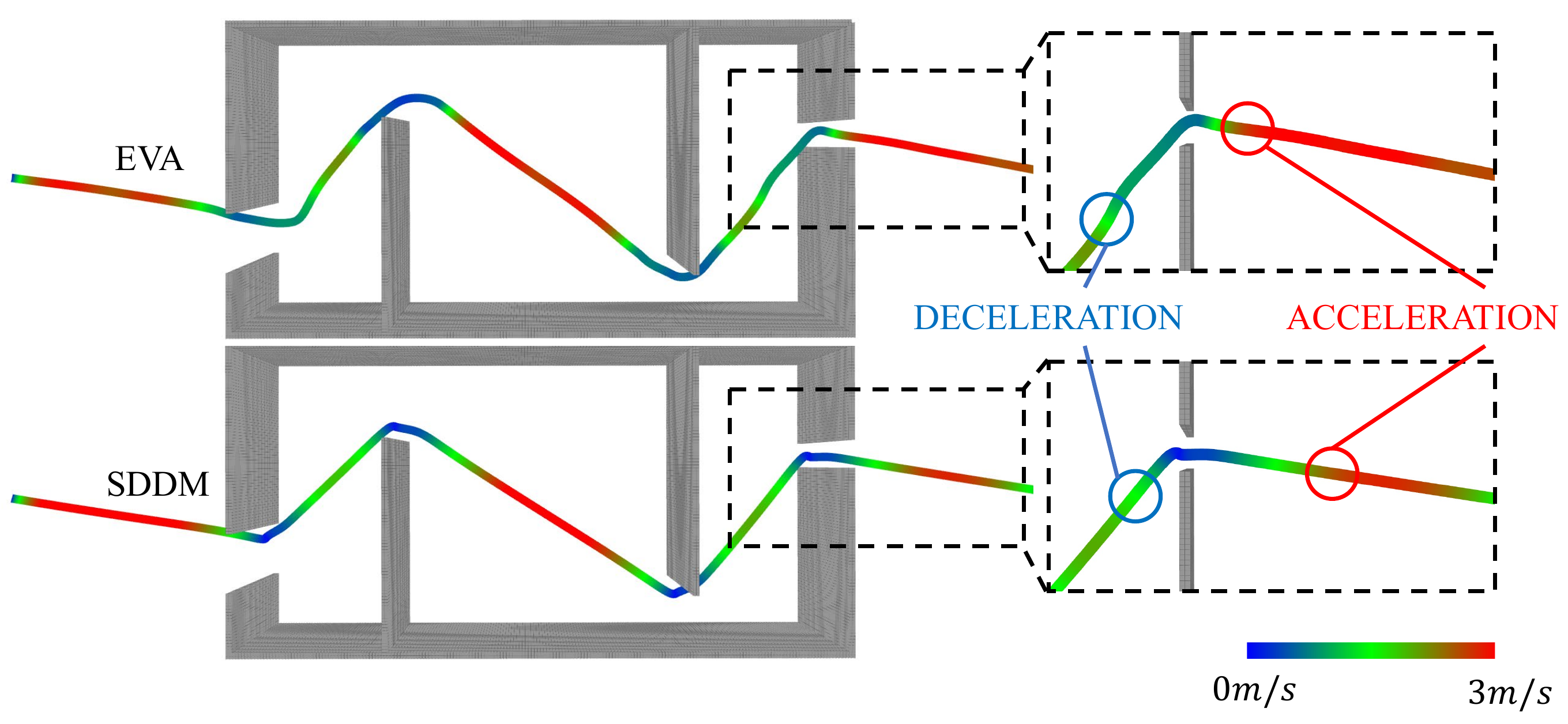}
	\end{center}
	\caption{ Velocity curve comparison of EVA and SDDM in the narrow gate map. Our method shows a more reasonable change in velocity when passing through a narrow gate in contrast to the zoom-in trajectories.
		\label{fig:benchmark}}
\end{figure}

We design a benchmark comparison to validate the effectiveness of EVA-planner. We compare EVA with SDDM ~\cite{li2020icra}, which also incorporates the velocity direction into the judgement of obstacles. However, this method is still conservative when the system flies away from the obstacles, as shown in Fig.\ref{fig:benchmark}. The performance of SDDM is adjusted to the best according to the parameters in~\cite{li2020icra}. The velocity limit is set as $3m/s$, and the same reference trajectories are generated. $40m\times5m$ random forest map and narrow gate map are used in tests. Comparisons of the running times are shown in Tab.\ref{tab:benchmark_time}. When the number of obstacles in the random forest map increases, the running time gap between EVA and SDDM gradually increases.

The benchmark shows that EVA-planner has better adaptability to the environment and can generate a more aggressive trajectory while ensuring safety.

\begin{table}[t]
	\caption{comparison of running times in different maps}
	\begin{tabular}{lccccc}
		\hline
		\hline
		\multirow{2}{*}{}Density$(obs./m^2)$& \multicolumn{4}{c}{Random forest map} & \multirow{2}{*}{Gate map} \\ 
		& 0.04       & 0.16      & 0.28      & 0.40      &                                  \\ \hline
		\multicolumn{1}{c}{SDDM time(s)} & 20.74   & 29.50   & 40.05   & 42.90   & 42.71                            \\
		\multicolumn{1}{c}{EVA time(s)}   & 19.28   & 25.02   & 31.04   & 33.01   & 32.99                            \\ \hline
		\hline
	\end{tabular}
\label{tab:benchmark_time}
\vspace{-1.0cm}
\end{table}

\begin{figure*}[t]
	\includegraphics[width=2.0\columnwidth]{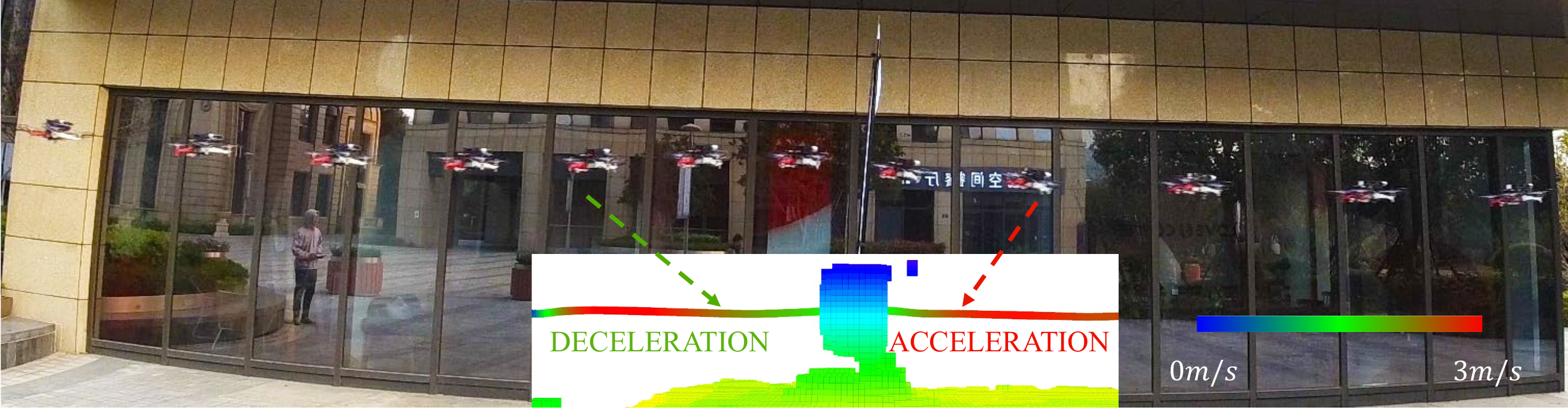}
	\caption{ Composite snapshots and executed trajectory of flying through a loop 
		\label{fig:through_circle}}
\end{figure*}

\begin{figure}[h]
	\begin{center}
		\includegraphics[width=0.85\columnwidth]{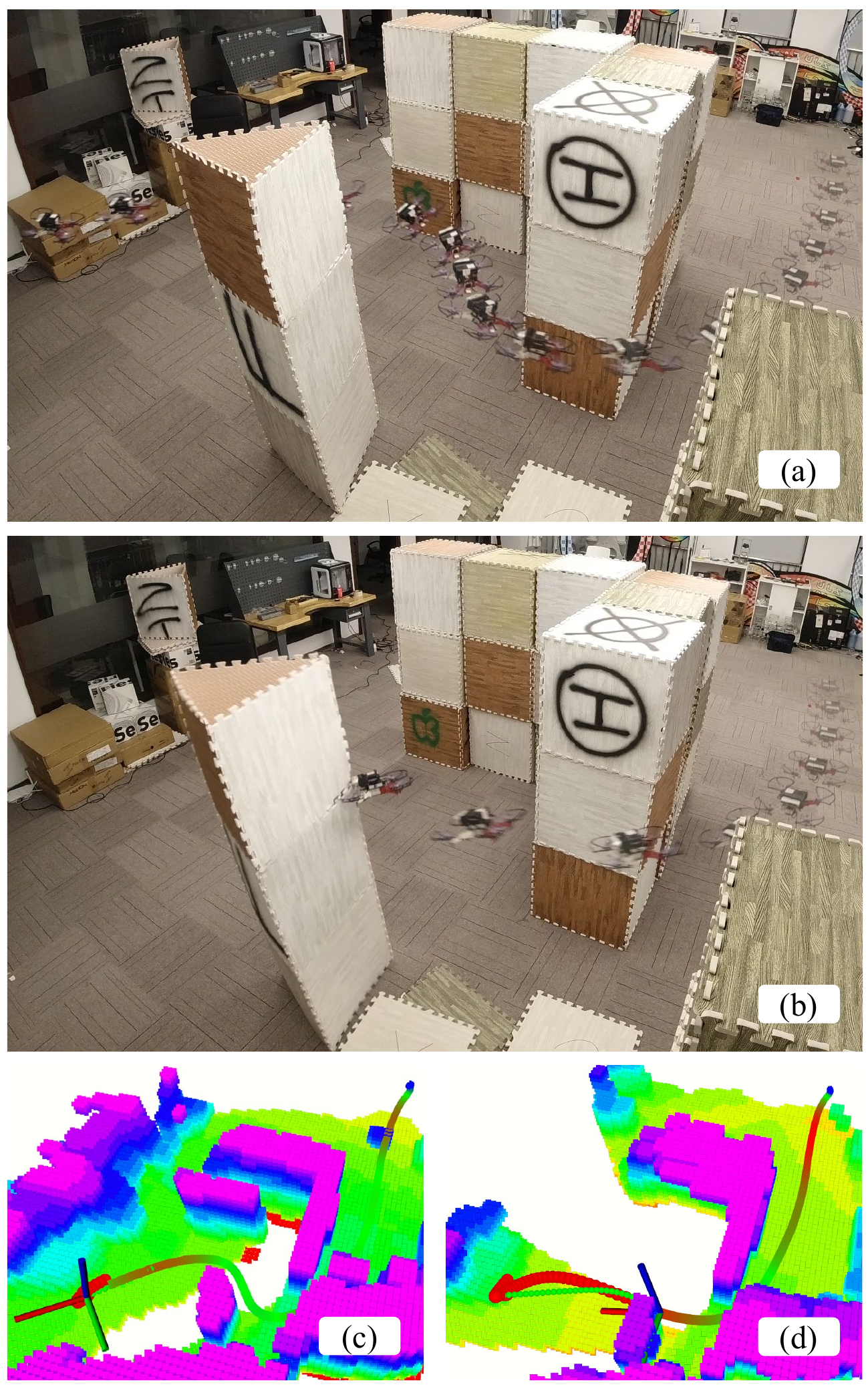}
	\end{center}
	\caption{ Composite snapshots and executed trajetcory of flying in a complex indoor environment with \& without EASA. The flight results in (a) and (c) are with EASA while (b) and (d) are without EASA.
		\label{fig:necessity_test}}
\end{figure}

\subsection{Real-world Test}
\label{sec:real_world}
We design several real-world experiments to validate the feasibility and robustness of our method. All calculations are run on the onboard computer in real-time.

In order to verify the characteristics of EVA-planner, we set up the quadrotor to go through a loop in the straight line, as shown in Fig.\ref{fig:through_circle}. Velocity limit is set as $3m/s$. Composite snapshots and executed trajectory show that a quadrotor slows down when it approaches the loop and accelerates as soon as it passes through it. The quadrotor flies in maximum velocity on the other times. This experiment verifies that EVA-planner can increase the aggressiveness of the flight trajectory while ensuring safety.

To verify the necessity of our method, we compare the flight results with \& without EASA in a complex indoor environment, as shown in Fig.\ref{fig:necessity_test}. We design a challenging scenario for UAV planning in response to the limited FOV of visual cameras. The quadrotor flies through a narrow gap and immediately turns to the right where an obstacle is placed in its path. In Fig.\ref{fig:necessity_test}b and Fig.\ref{fig:necessity_test}d, trajectory without EASA performs unsafe aggressiveness when it flies through the narrow gap. This behavior causes the quadrotor to have no time to avoid the sudden obstacle. By contrast, from Fig.\ref{fig:necessity_test}b and Fig.\ref{fig:necessity_test}d, the quadrotor with EASA performs conservative velocity when passing the dangerous area, which makes the system have enough time to deal with the sudden danger. We repeat the experiments with EASA seven times with a success rate of 100\%. However, two out of three experiments without EASA fail and hit the obstacles.

In the outdoor experiment, the quadrotor flies from an open area into the bushes, then out of the bushes into another open area, as shown in Fig.\ref{fig:real}. Although the bushes are rugged and leafy, the quadrotor can still carry out adaptive deceleration while flying into the dangerous area and adaptive acceleration while flying out of the dangerous area. As validated in the attatched video, we repeat ten flights, and all are successful. This experiment verifies the robustness of our algorithm in the complex unknown environment.

\section{Conclusions and Future Work}
\label{sec:conclusion}
In this paper, we propose a novel environmental adaptive safety aware method according to the gradient of ESDF and the velocity. Compared with the benchmark method, this method is more reasonable in judging the degree of danger. Therefore it can plan a more aggressive trajectory without affecting safety. 

To apply this method into the trajectory generation, we design a unified multi-layered planning framework to generate smooth, adaptively aggressive, safe, and dynamical feasible local trajectories. Extensive simulations and real-world experiments validate the robustness and effectiveness of our planning framework. In the future, we plan to generate environmental adaptive trajectories in complex dynamic environments and intend to challenge our quadrotor system in more extreme situations.

\bibliography{icra2021quanlun}
\end{document}